\documentclass[%
 aip,
 amsmath,amssymb,
preprint,%
]{revtex4-2}

\usepackage{graphicx}
\usepackage{dcolumn}
\usepackage{bm}

\usepackage[utf8]{inputenc}
\usepackage[T1]{fontenc}
\usepackage{etoolbox}

\graphicspath{{./eps/}{./epstopdf_converted/}}

\newcommand*\G {\mathcal{G}}

\newcommand*\D {\mathcal{D}}
\newcommand*\F {\mathcal{F}}
\newcommand*\R {\mathbb{R} } 

\newcommand* \disp {\phi}

\newcommand*\z {\varepsilon}

\newcommand*\kk {\kappa}

\newcommand* \xHat    {\hat{x}}
\newcommand* \tHat    {\hat{t}}
\newcommand* \dispHat {\psi}

\newcommand* \RevisionDelete[1] {}

\makeatletter
\def\@email#1#2{%
 \endgroup
 \patchcmd{\titleblock@produce}
  {\frontmatter@RRAPformat}
  {\frontmatter@RRAPformat{\produce@RRAP{*#1\href{mailto:#2}{#2}}}\frontmatter@RRAPformat}
  {}{}
}%
\makeatother

\begin{document}



	\title{ Learning Flame Evolution Operator under Hybrid Darrieus Landau and Diffusive Thermal Instability  }
	\author{Rixin Yu }
	\email{rixin.yu@energy.lth.se}
	\affiliation{ 
	Department of Energy Sciences, Lund University, 22100 Lund, Sweden
	}%
	
	\author{Erdzan Hodzic}
	\affiliation{%
	Department of Manufacturing Processes, RISE Research Institutes of Sweden, 553 22 Jonkoping, Sweden 
	}%

	\author{Karl-Johan Nogenmyr}
	\affiliation{Siemens Energy AB, 612 31 Finsp{\aa}ng, Sweden}
	
	\date{\today}
	
	\begin{abstract}
Recent advancements in the integration of artificial intelligence (AI) and machine learning (ML) with physical sciences have led to significant progress in addressing complex phenomena governed by nonlinear partial differential equations (PDE). This paper explores the application of novel operator learning methodologies to unravel the intricate dynamics of flame instability, particularly focusing on hybrid instabilities arising from the coexistence of Darrieus-Landau (DL) and Diffusive-Thermal (DT) mechanisms. Training datasets encompass a wide range of parameter configurations, enabling the learning of parametric solution advancement operators using techniques such as parametric Fourier Neural Operator (pFNO), and parametric convolutional neural networks (pCNN). Results demonstrate the efficacy of these methods in accurately predicting short-term and long-term flame evolution across diverse parameter regimes, capturing the characteristic behaviors of pure and blended instabilities. Comparative analyses reveal pFNO as the most accurate model for learning short-term solutions, while all models exhibit robust performance in capturing the nuanced dynamics of flame evolution. This research contributes to the development of robust modeling frameworks for understanding and controlling complex physical processes governed by nonlinear PDE.
 	\end{abstract}
	
	\maketitle



\section{Introduction \label{sec:intro}} 

In recent years, the integration of artificial intelligence (AI) and machine learning (ML) with the natural sciences and physical engineering has led to significant advancements, particularly in addressing the complexities of nonlinear partial differential equations (PDE). These equations are fundamental in understanding various physical phenomena, ranging from turbulent fluid dynamics to complicate physico-chemical processes. Within the domain of nonlinear PDE systems lies a rich tapestry of intricate dynamics, including instabilities, multiscale interactions, and chaotic behaviors. To enhance predictive capabilities and design robust control strategies in engineering applications, computational methods are indispensable. These methods, often in the form of numerical solvers, enable the accurate simulation of PDE solutions across spatial and temporal domains. Implicit in these solvers is the concept of the functional-mapping operator, which could iteratively advances the PDE solution functions in time, providing a pathway to explore the evolution of physical systems over extended durations. A distinctive class of machine learning methods has emerged, capable of learning and replicating the behavior of these PDE operators.

Recent advancements have seen the proliferation of operator learning methods, each offering unique insights and capabilities. Early efforts in this domain drew inspiration from deep convolutional neural networks (CNN)\cite{CNN1,CNN2,CNN3,CNN4,CNN5,UNet,ConvPDE}, employing techniques from computer vision. These CNN-based approaches parameterize the PDE operator in a finite-dimensional space, enabling the mapping of discrete functions onto image-like representations. Building upon this foundation, recent strides have witnessed the development of neural operator methods\cite{GraphKenerlNetwork, kovachki2021neural} capable of learning operators of infinite dimensionality. Notable examples include Deep Operator Network\cite{DeepONet} and the Fourier Neural Operator (FNO)\cite{FNO2020}, both demonstrating remarkable proficiency across a diverse array of benchmark problems\cite{DeepONet_Nature, DeepONet_FNO_cmp}. Furthermore, recent advancements have extended neural operators by amalgamating concepts from wavelet methods \cite{gupta2021multiwavelet, tripura2023wavelet} and adapting approaches for complex domains \cite{chen2023laplace}.

In our recent investigations \cite{Yu2023,YuH2024}, we delved into the intricate dynamics of flame instability and nonlinear evolution, a canonical problem with profound implications for combustion science. Flames can undergo destabilization due to intrinsic instabilities, including the hydrodynamic Darrieus-Laudau (DL) mechanism \cite{DARRIEUS1938UNPB,landau1988theory} attributed to density gradients across a flame, and the Diffusive-Thermal (DT) mechanism \cite{zeldovich1944selected, sivashinsky1977diffusional}  driven by heat and reactant diffusion disparities. Our previous work \cite{Yu2023} primarily focused on DL flames, scrutinizing the evolution of unstable flame fronts within periodic channels of varying widths. Under DL instability, an initially planar flame morphs into a steady curved front; as the channel width increases, the curved front becomes sensitive to random noise, and small wrinkles start to emerge. At sufficiently large channels, DL flames give rise to complicated fractal fronts characterized by hierarchical cascading of cellular structures \cite{YBB15PRE}.

The nonlinear evolution of DT flame development can be modeled by the Michelson-Sivashinsky equation \cite{michelson1977nonlinear}, while a more accurate but computationally expensive approach involves direct numerical simulation (DNS) of Navier-Stokes equations. Utilizing these two approaches to generate training datasets, our investigations  \cite{Yu2023} demonstrated that both CNN and FNO could effectively capture the evolution of DL flames, with FNO exhibiting superior performance in modeling complex flame geometries over longer durations. Subsequently, we embarked on developing parameterized learning methodologies capable of encapsulating dynamics across diverse parameter regimes within a single network framework. Through the introduction of pCNN and pFNO models \cite{YuH2024}, we demonstrated their efficacy in replicating the behavior of DL flames across varying channel widths. Additionally, our methods have shown success in learning the parametric solutions of the Kuramoto-Sivashinsky equation \cite{kuramoto1978diffusion}, which models unstable flame evolution due to the DT mechanism. However, a challenge remains in mitigating the tendency of these models to overestimate noise effects.

In this paper, we extend our research horizon to encompass the complexities arising from hybrid instabilities, specifically those arising from the coexistence of DL and DT mechanisms. These hybrid systems pose new challenges, as they embody a rich spectrum of behaviors stemming from the interplay of distinct instability modes. Leveraging our novel operator learning methodologies, we aim to unravel the nuanced dynamics underlying such hybrid instabilities, shedding light on their short-term evolution and long-term statistical properties. Furthermore, our endeavor holds promise for the development of robust modeling frameworks capable of capturing the intricate dynamics of real-world flame evolution scenarios.

The paper is organized as follows: first, we describe the problem setup for learning PDE operators, followed by brief descriptions of the two parametric learning methods to be used in this work. These methods will be compared in the context of learning parametric-dependent solution time-advance operators for the Sivashinsky Equation \cite{SivaEq}, which models unstable front evolution due to hybrid mechanisms of flame instability. Finally, we provide a summary and conclusion.

\section{ Problem Setup for Learning PDE Operators \label{sec:OLmethods} }

In this section, we delineate the problem setup for learning a parametric PDE operator, along with a description of recurrent training methods.

Consider a system governed by PDE, typically involving multiple functions and mappings between them. Our focus here is on a parametric operator mapping, denoted as
\begin{equation}
\hat{\mathcal{G}}: \mathcal{V} \times \mathbb{R}^{d_\gamma} \to \mathcal{V}'
;
(v(x), \gamma) \mapsto v'(x')
\end{equation}
where $\gamma \in \mathbb{R}^{d_\gamma}$ represents a set of parameters. The input function $v(x)$, where $x \in\D$, resides in a functional space $\mathcal{V}(\D,\mathbb{R}^{d_v})$, with domain $\D \subset \mathbb{R}^d $ and codomain $\mathbb{R}^{d_v}$, while the output function $v'(x')$, where $x' \in\D'$, belongs to another functional space $\mathcal{V}'(\D',\mathbb{R}^{d_v'})$ with domain $\D' \subset \mathbb{R}^{d'} $ and codomain $\mathbb{R}^{d_v'}$.

Our primary interest lies in the solution time advancement operator with parametric dependence, given by
\begin{equation}
\hat{\mathcal{G}} : (\disp(x,\bar{t}), \gamma) \mapsto \disp(x;\bar{t}+1)
\label{eq:G_operator}
\end{equation}
where $\disp(x;\bar{t})$ denotes the solution to a PDE under parameters $\gamma$, and $\bar{t} = t/\Delta_t$ represents normalized time with a small time increment $\Delta_t$. For simplicity, we assume identical domain and codomain for both input and output functions, i.e., $\D' = \D$, $\mathcal{V'} = \mathcal{V}$, $d'=d $, and $d'_v=d_v$, with periodic boundary conditions on $\D$.

To approximate the mapping $\hat{\mathcal{G}}$ using neural network methods, let $\Theta$ denote the space of trainable parameters in the network. A neural network can be defined as
\begin{equation}
 \mathcal{G}: \mathcal{V}\times \mathbb{R}^{d_\gamma} 
\times \Theta \to \mathcal{V}' \\ 
\text{ or equivalently }
 \mathcal{G}_{\theta}: \mathcal{V} \times \mathbb{R}^{d_\gamma}
 \to \mathcal{V}', \theta \in  \Theta. 
\end{equation}
where $\Theta$ represents the space of network parameters. Training the neural network involves finding an optimal choice of parameters $\theta^* \in \Theta$ such that $\mathcal{G}_{\theta^*}$ approximates $\hat{\mathcal{G}}$.

Starting with an initial solution function $\phi(x;t_0)$ under fixed parameter values $\gamma$, the recurrent application of the operator $\mathcal{G}_{\theta, \gamma}:=\mathcal{G}_{\theta}(\cdot, \gamma) $ can roll out predicted solutions of arbitrary length by iteratively updating the input function with its output from the previous prediction.
Note, while the learned operator is expected to make accurate short-term predictions; its long-term prediction might be allowed to deviate if the ground truth PDE admits chaotic solutions. On the other hand, it is still desirable that the learned operator can reproduce the correct statistics in the long-term solutions.

Following previous studies \cite{Yu2023, YuH2024}, our training approach adopts a one-to-many setup where the recurrent network is trained to make multiple successive predictions from a single input function. 
Such setup ensures numerical stability in the learned solution advancement operator, a crucial consideration highlighted in the prior work \cite{Yu2023,YuH2024}.
More specifically, let 
$\left\{ 
v_{j}, (\hat{\mathcal{G}}_{\gamma_i}^1 v_j, \hat{\mathcal{G}}_{\gamma_i}^2 v_j,..., \hat{\mathcal{G}}_{\gamma_i}^n v_j ) 
\right\}_{j=1,i=1}^{ j=Z',i=Z''}$ 
be a total ($Z'\times Z''$)-number of training data arranged as input/output pairs in the 1-to-$n$ manner, and, an operator with a superscript $n$ denotes its repeated application $n$ times, e.g. $ \hat{\mathcal{G}}^n_\gamma := \hat{\mathcal{G}}_\gamma \circ ... \circ \hat{\mathcal{G}}_\gamma $.
Training a network $\mathcal{G}_\theta$ to approximate $\hat{\mathcal{G}}$ then becomes a minimization task
\begin{equation} 
  \min_{\theta\in \Theta} \mathbb{E}_{ v \sim \chi', \gamma \sim \chi'' }
 \left [ 
C (
  ( \mathcal{G}_{\theta,\gamma}^1 v,..., \mathcal{G}_{\theta,\gamma}^n v ) 
, 
(\hat{\mathcal{G}}_\gamma^1 v, ,..., \hat{\mathcal{G}}_\gamma^n v )
) 
\right ]
\label{eq:opt_1-to-n}
\end{equation}
where $v \sim \chi'$ and $\gamma \sim \chi''$ are randomly drawn according to independent probability measures of $\chi'$ and $\chi''$ respectively.
The cost function $C : \mathcal{V}^n \times \mathcal{V}^n \to \R$ is set to the relative mean square (L2) error of $C(x,y)=||x-y||_2/||y||_2$,
here $\mathcal{V}^n$ abbreviates the Cartesian product of $n$ copies of $\mathcal{V}$.

\section{Parametric Operator Learning Methods \label{sec:networks} }

In this section, we present a concise overview of two methods capable of learning the parametric operator $\hat{\G}$. Further details about these methods can be found in paper \cite{YuH2024}.

\subsection{Parametric Convolutional Neural Network (pCNN)}

\begin{figure*}
	\centerline{
		\includegraphics[width=1\linewidth]{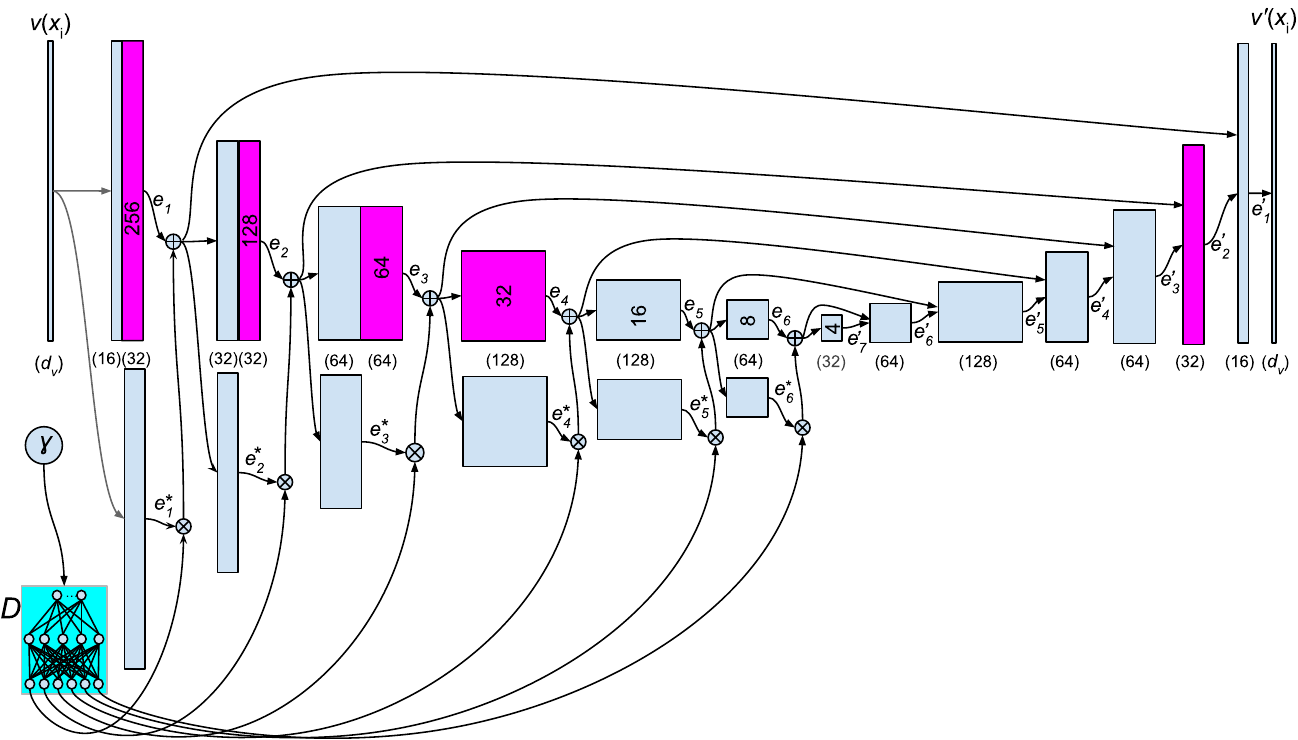}
	}
	\caption
	{
		\label{fig:CNN}
The parametric CNN model adopted in this study is demonstrated for input function $v(x_i)$ discretized at 1D mesh of 256 points, with $L=6$ levels of encoding and decoding. Standard convolution layers are represented by gray rectangles, while Inception layers are depicted in magenta. The output data channels $c_l$ for each convolution layer are indicated within brackets.
	}
\end{figure*}

The operator  $\mathcal{G}_{\theta,\gamma} $ can be regarded as an image-to-image map when applied for the temporal advancement of discretized solutions. Deep Convolutional Neural Networks (CNN) have demonstrated effectiveness in image learning tasks. The network architecture suitable for learning operators resembles a convolutional auto-encoder similar to that in U-Net \cite{UNet} and ConvPDE \cite{ConvPDE}. This network comprises an encoder block and a decoder block, with input data passing through a series of transformations via convolutional layers. Additionally, the method incorporates side networks to handle additional parameter inputs. The pCNN model  is outlined in Fig. \ref{fig:CNN}.

Let $e^+_0$ denote the input function $v(x_j)$ represented on an $x$-mesh. The encoder block follows an iterative update procedure:   $ e^+_{l} \mapsto (e_{l+1},e^*_{l+1} ) \mapsto e^+_{l+1} $. 
This iteration occurs over the level sequence $l = 0,1,...,L-1$. 
Denote the last decoding output as $e'_L=e^+_L$, a subsequent decoding procedure is applied $ (e'_{l+1},e^+_{l}) \mapsto e'_{l} $ through reversing level $l$.

Here, $e_l, e^*_l, e^+_l , e'_l \in \mathbb{R}^{c_l\times N_l}$ represent four data sequences, each with $c_l$ channels and size of $N_{l}$. The data size is halved as $N_{l+1}=N_{l}/2$ for $l\geq 1$. The first-stage encoder contains two sub-maps of $e^+_{l} \mapsto e_{l+1}$ and $e^+_{l} \mapsto e^*_{l+1}$, both are implemented using vanilla-stacked convolution layers (with a filter size of 3, stride 1, periodic padding, and ReLU activation). Some layers are replaced by Inception layers for improved performance. Additionally, a size-2 max-pooling layer is prepended to halve the image size for $l\geq1$. The second-stage encoder map is implemented as $e^+_{l+1} = e_{l+1} +  e^*_{l+1}\cdot D_{l}(\gamma)$.
Here, $D_l$ is a simple function (a two-layer perceptron) that converts the PDE parameters $\gamma$ into a scaling ratio. The decoder update $(e'_{l+1},e^+_{l})\mapsto e'_{l}$ involves concatenating $e'_{l+1}$ (but up-sample it to double its size) with $e^+_{l}$ along the channel dimension. The final output is obtained as $v'(x_j)=e'_1$.

\subsection{Parametric Fourier Neural Operator (pFNO) \label{sec:FNO} }

The parametric Fourier Neural Operator (pFNO) \cite{YuH2024} was developed based on the original FNO  method \cite{FNO2020}, wherein learning for the infinite-dimensional operator is achieved by parameterizing the integral kernel operators in Fourier Space. The pFNO adopts an architecture of maps-composition as $ \mathcal{G} = \mathcal{Q} \circ \mathcal{H}_L \circ ... \circ \mathcal{H}_1 \circ \mathcal{P}\circ \mathcal{C} $, comprising a concatenation map $\mathcal{C}$ , a lifting map $\mathcal{P}$, a sequence of hidden maps $\mathcal{H}_l$ for $l=1,2, \ldots, L$, and a projection map $\mathcal{Q}$.

The first map $\mathcal{C}: \mathcal{V} \times \mathbb{R}^{d_\gamma} \to \mathcal{V}^c(\mathcal{D}; \mathbb{R}^{d_v + d_\gamma} )  ; (v(x), \gamma) \mapsto v^c(x) $ simply concatenates the parameters $\gamma$ to the co-dimension of input function $v(x)$, yielding  $v^c(x)$. 
The second map $ \mathcal{P}: \mathcal{V}^c  \to \mathcal{V}^* ; v^c(x) \mapsto \varepsilon_{0} (x) $, lifts the input to a higher-dimensional functional space $\mathcal{V}^*:=\mathcal{V}^*(\mathcal{D}; \mathbb{R}^{d_\varepsilon})$ with $d_\varepsilon > d_v+d_\gamma $. 
The subsequent hidden maps $\mathcal{H}_l: \mathcal{V}^* \times \mathbb{R}^{d_\gamma} \to \mathcal{V}^* : (\z_{l-1}, \gamma) \mapsto \z_{l}$ act sequentially to update $\z_0 \mapsto \z_1 \mapsto \ldots \mapsto \z_L$ for all $\z_l  \in \mathcal{V}^*$. Finally, the map $\mathcal{Q}: \mathcal{V}^* \to \mathcal{V}'; \z_L(x) \mapsto v'(x)$ projects back to low dimension functional space, finally yielding $v'(x)$.

Both $\mathcal{P}$ and $\mathcal{Q}$ are implemented using simple multilayer perceptrons(MLP). 
The hidden maps $\mathcal{H}_{l+1}$ are implemented as parametric Fourier layers:
\begin{equation}
 \varepsilon_{l+1} =  \sigma\left( W_{l} \z_l  + b_l +  \mathcal{F}^{-1} \{ \mathfrak{R}^*_l( \mathcal{F} \{ \z_l \}, \gamma )  \} \right)
\label{eq:pFNO}
\end{equation}
where $W_l \in \mathbb{R}^{d_\varepsilon \times d_\varepsilon}$ and $b_l \in \mathbb{R}^{d_\varepsilon}$ are learnable weights and biases respectively, and $\sigma$ is a ReLU activation function. Here, $\mathcal{F}$ and $\mathcal{F}^{-1}$ represent the Fourier Transform and its inverse respectively.
The  function $\mathfrak{R}^*_l: \mathbb{C}^{\kappa^{max} \times  d_\z} \times \mathbb{R}^{d_\gamma}  \to \mathbb{C}^{\kappa^{max} \times d_\z } $ acts on the truncated Fourier modes, transforming them as :
\begin{eqnarray}
 \mathfrak{R}^*_l (\F\{\z\} , \gamma )_{\kappa,i} =   
     \sum_{j=1}^{d_\z} [ (R_l)_{\kappa,i,j}   +(R^*_l)_{\kappa,i,j}   D_l^*(\gamma)_{\kappa}  ] \F \{\z\}_{\kappa,j} , \nonumber \\
\kappa=0,1,...,\kappa^{max} \text{ and }  i=1,...,d_\z
\label{eq:Rstar}
\end{eqnarray}
where $R_l, R^*_l \in  \mathbb{C}^{\kappa^{max} \times d_\z \times d_\z}$ are two learnable weight tensors, and  $D^*_l: \R^{d_\gamma}  \to  \R^{\kappa^{max}} $ is a function  converting the parameters $\gamma$ into $\kappa^{max}$-number of scaling ratios. 
This function consists of  a two-stage map $\gamma \mapsto D_l(\gamma) \mapsto D^*_l(\gamma)$, with   $D_l(\gamma) \in \mathbb{R}^{N_D}$  outputting $N_D$ scaling ratios and implemented as a MLP. The second map hierarchically redistribute these ratio across the wave numbers. In one dimension($d=1$), the distribution map reads:
$   D^*_l(\gamma) _\kappa =    D_l(\gamma) _i$  for $\kappa \in  ( \frac{ \kappa^{max}}{2^{i+1}} , \frac{ \kappa^{max}}{2^{i}} ] $  at $i =0,..,N_D-2$,  and, for $ \kappa  \in  (0, \frac{ \kappa^{max}}{2^{N_D-1}}] $   at $ i =N_D-1$.

One might observes that we can deactivate the second weight tensor $R^*_l$ in Eq. \eqref{eq:Rstar} by enforcing the map $D^*_l(\gamma)$ to output only zeros. This modification still enables learning of the parameter operator due to the concatenation map $\mathcal{C}$. 
Such a modified method can viewed as a simple tweak to the baseline method of FNO \cite{FNO2020} and will be referred as pFNO* in later section.

\section{ Numerical experiments and result discussions \label{sec:numericalexp} }

In this section, we employ the pFNO and pCNN methods to learn flame evolution under hybrid instabilities arising from both Darrieus-Landau (DL) \cite{DARRIEUS1938UNPB,landau1988theory} and Diffusive-Thermal (DT) \cite{zeldovich1944selected, sivashinsky1977diffusional} mechanisms. 
The dynamics of such unstable flame development are encapsulated by the Sivashinsky equation \cite{SivaEq}. 
To facilitate parametric learning, we begin by reformulating the Sivashinsky equation, introducing two parameters that enable straightforward specification for blending the two instabilities and controlling the largest unstable wave numbers. By sampling across these parameters, we construct an extensive training dataset covering a range of relevant scenarios subjected to different  DL/DT mixing. Subsequently, we present the results and compare the performance of the different methods in learning these hybrid instabilities.

\subsection{ Governing equations \label{sec:1d}}

Consider modeling the unstable development of a statistically planar flame front. Let $\tHat$ denote time and $\xHat$ represent the spatial coordinate along the normal direction to flame propagation. Introduce a displacement function $\dispHat(\xHat,\tHat): \R \times \R \to \R$  describing the stream-wise coordinate of a flame front undergoing intrinsic flame instabilities.
Such evolution can be modeled by the Sivashinsky equation \cite{SivaEq}:
\begin{eqnarray} 
 \dispHat _{\tHat}  + \frac{1}{2} \left( \dispHat _{\xHat}  \right)^2 =
-  4(1 + \text{Le}^*)^2  \dispHat _{\xHat\xHat\xHat\xHat}
    -  \text{Le}^*  \dispHat _{\xHat\xHat} 
     + (1-\Omega)  \Gamma( \dispHat )  
 \label{eq:MKS_org}
\end{eqnarray}
where $\Gamma: \dispHat \mapsto -\mathcal{H}(\dispHat_{\xHat})$ is a linear singular non-local operator defined using the Hilbert transform $\mathcal{H}$, or equivalently written as $\Gamma: \dispHat \mapsto \F^{-1} (|\kk| \F_\kk (\dispHat))$
using the spatial Fourier transform $\F_\kappa(\dispHat )$ and its inverse $\F^{-1}$.

In Eq. \eqref{eq:MKS_org},  $\Omega$ is the density ratio between burned product and fresh reactant;  $\text{Le}^*$ is a ratio (positive or negative) depending on the Lewis number of deficient reactant and another critical Lewis number.
Introduce three constants ($a,b,c$) for variable transformation on time $t= a^2 \tHat $, space $x= b^{-2} \xHat$ and the displacement function $\dispHat(\xHat,\tHat) = c^2 \disp(x,t)$, then equation \eqref{eq:MKS_org} can be rewritten
\begin{eqnarray}
 \frac{1}{\tau}  \disp_t  
+ \frac{1}{2 \beta^2 } ( \disp_x  )^2 =
-  \frac{ \mu }{ \beta^4}   \disp_{xxxx}
    - \frac{ \nu }{ \beta^2}   \disp_{xx}
     +  \frac{\rho}{\beta} \Gamma(\disp)  
 \label{eq:MKS}
\end{eqnarray}
with $\beta =bc^{-1}$, $\nu=\text{Le}^*/c^2$, $\rho=(1-\Omega)bc $, 
$\mu = 4(1+\text{Le}^*)^2 b^{-2} c^{-4}$ and $\tau =a^{-2}b^{-2}$. 

In this work we consider the flame front solution $\disp(x,t)$ of Eq. \eqref{eq:MKS} in a channel domain subjected to periodic boundary condition, i.e. $x\in \D=(-\pi, \pi]$. 
One might notice that Eq. \eqref{eq:MKS} admits a zero equilibrium solution being a flat flame (i.e. $\phi^*(x,t)=0$), a perturbation analysis around this zero solution yields a linear dispersion relation 
\begin{eqnarray}
\frac{ \omega (\kappa) }{\tau } =  
       -\mu   \left(\frac{\kappa}{\beta} \right)^4 
        +  \nu  \left(\frac{\kappa}{\beta} \right)^2 
         +  \rho  \left(\frac{\kappa}{\beta} \right),  \;  \forall \kappa = 0,1,2,...  
\label{eq:dRel_ana}
\end{eqnarray}
with the perturbed solution being $\phi(x,t) = \sum_{\kappa} \hat{\phi}_\kappa(t) e^{i \kappa x } +  \hat{\phi}^*_\kappa(t) e^{-i \kappa x }  $ (superscript * denotes complex conjugate) and the Fourier mode of perturbation evolving as $\hat{\phi}_{\kappa}(t) \approx \hat{\phi}_{\kappa}(0) \cdot e^{ \omega (\kappa ) \cdot t }$. 

Equations \eqref{eq:MKS} and \eqref{eq:dRel_ana} present a straightforward approach to hybridize two flame instabilities of DT and DL mechanisms. This strategy is accomplished by specifying two parameters, $\rho$ and $\beta$, while the remaining parameters ($\mu$, $\nu$, and $\tau$) can be determined by additional constraints outlined below. Initially, the parameter $\rho$ (in between 0 and 1) is defined to allow for the continuous blending of these two instabilities. When $\rho=1$, the Sivashinsky equation \eqref{eq:MKS} yields a pure DL instability described by the Michelson-Sivashinsky (MS) equation \cite{michelson1977nonlinear}:
\begin{eqnarray}
\frac{1}{\tau}\disp_t + \frac{1}{2 \beta^2} (\disp_x )^2 =\frac{ 1}{\beta^2} \disp_{xx} +\frac{1}{\beta} \Gamma(\disp) \label{eq:MS}
\end{eqnarray}
whereas, at the other end ($\rho=0$), it recovers the pure DT instability as described by the Kuramoto-Sivashinsky (KS) equation \cite{kuramoto1978diffusion}:
\begin{eqnarray}
\frac{1}{\tau} \disp_t + \frac{1}{2 \beta^2} (\disp_x )^2 = -\frac{1}{\beta^2} \disp_{xx} - \frac{1}{\beta^4} \disp_{xxxx} . \label{eq:KS}
\end{eqnarray}

\begin{figure*}
	\centerline{		
\includegraphics[width=1\linewidth]{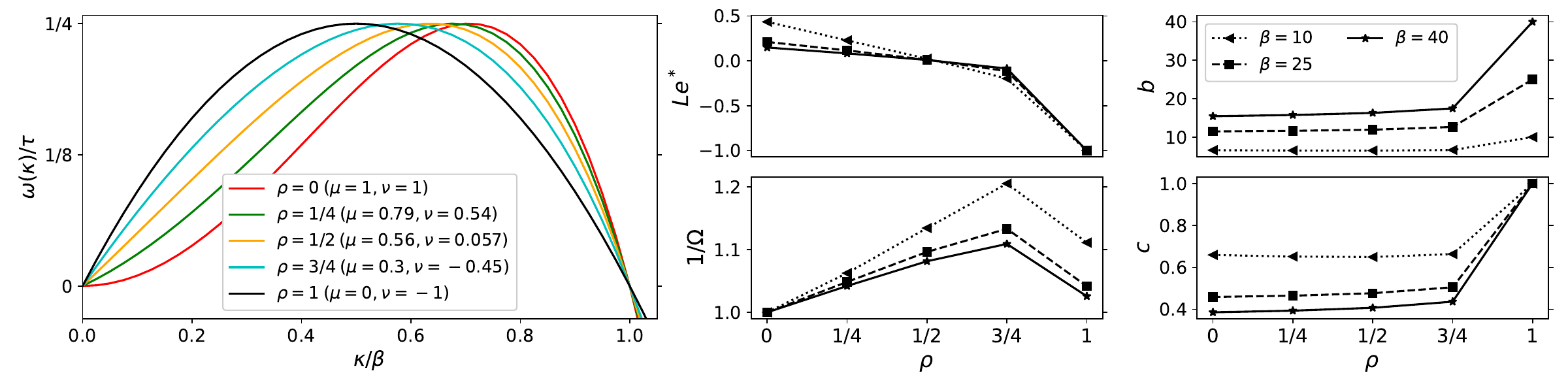}   
}
	\caption{ Dispersion relations (left) and relevant parameter values at prescribed values of $\rho$ and $\beta$.		\label{fig:dRel_ana}	}
\end{figure*}

Secondly, the parameter $\beta$ is determined as the largest value for which the dispersion relation of Eq. \eqref{eq:dRel_ana} equals zero (i.e. $\omega(\beta)=0$). This definition yields $\nu = \mu - \rho$. Consequently, we can prescribe $\beta$ to establish the largest unstable wave number. To mitigate variability in the remaining parameters, a third constraint is imposed: the maximum value of $\omega(\kappa)$ over the interval $0<\kappa<\beta$ must be $1/4$. Furthermore, $\tau = \rho \beta/10+ (1-\rho)$ is employed to better accommodate the timescales attributed to the various hybrid instabilities. This strategy allows for the determination of all remaining parameters given the values of $\rho$ and $\beta$. This is illustrated in Figure \ref{fig:dRel_ana}, which presents dispersion relation plots and associated parameters.

Before proceeding further, it may be worthwhile to mention a few well-known results. The KS equation \eqref{eq:KS} is often utilized as a benchmark example for PDE learning studies and is renowned for exhibiting chaotic solutions at large $\beta$. On the other hand, the MS equation \eqref{eq:MS}, although less familiar outside the flame instability community, can be precisely solved using a pole-decomposition technique \cite{Thual_Frisch_Henon_poledecomp}, transforming it into a set of ODEs with finite freedoms. Moreover, at large $\beta$, the MS equation admits a stable solution in the form of a giant cusp front. However, at smaller $\beta$, the equation becomes susceptible to noise, resulting in unstable solutions characterized by persistent small wrinkles atop a giant cusp. Additional details about known theory can be found in references \cite{Vaynblat_matalon_polestability1, Vaynblat_matalon_polestability2, Olami_noise, denet2006stationary, Kupervasser_pole_book, Karlin2002cellular, Creta2020propagation, CRETA2011INST,YBB15PRE}.

\subsection{Training dataset \label{sec:dataset}}

Equation \eqref{eq:MKS} is tackled using a pseudo-spectral approach combined with a Runge-Kutta (4,5) time integration method. All solutions are computed on a uniformly spaced 1D mesh consisting of 256 points.
Training datasets are generated for a total of 15 parametric configuration tuples $(\rho, \beta) $, formed as the Cartesian product of three values for $\beta$ in the range $[10, 25, 40]$ and five values for $\rho$ in the range $[0, 1/4, 1/2, 3/4, 1]$.  
For each of the fifteen parametric configurations, we generate 250 sequences of short-duration solutions, as well as a single sequence of long-duration solutions. 
Each short solution sequence spans a time duration of $0 \leq t \leq 75$ and contains 500 consecutive solutions separated by a time interval of $\Delta_t = 0.15$. 
Additionally, each sequence starts from random initial conditions $\disp_0(x)$ sampled from a uniform distribution over the range $[0, 0.03]$. 
The long sequence covers a time duration of $0 \leq t \leq 18,750$ and comprises 125,000 consecutive solutions outputted at the same interval $\Delta_t$. 
A validation dataset is similarly created for all fifteen parameter tuples, but it contains only 10 percent of the data present in the training dataset.

\subsection{ Result analysis }

\begin{table}
	\caption{  Relative $L2$ train/validation errors for all operator learning networks \label{Table1} }  
	\centerline{
		\begin{tabular}{|c|r|c|c|c|c}  
			\hline
			   Model &  Parameter configurations $(\beta,\rho)$ &	Train $L2$	& Valid. $L2$	\\
			\hline
			pFNO*  &  $[10,25,40] \times [0,1/4,1/2,3/4,1]$    &    0.0098  & 0.010    \\ 
			pFNO &  $[10,25,40] \times [0,1/4,1/2,3/4,1]$   &  0.0071    &  0.0073\\ 
  			\hline
			pCNN   &  $[10,25,40] \times [0,1/4,1/2,3/4,1]$    &  0.036   &  0.037 \\ 
  			\hline
			pCNN10   &  $[10] \times [0,1/4,1/2,3/4,1]$    &  0.011   &  0.012 \\ 
			pCNN40  &  $[40] \times [0,1/4,1/2,3/4,1]$    &  0.022   &  0.022 \\ 
  			\hline
		\end{tabular}
	}
\end{table}

The training datasets described in the previous section are utilized to train parametric solution advancement operators, denoted as $\hat{\mathcal{G}}_{(\gamma)} : \phi(x;t)   \mapsto \phi(x;t+\Delta_t)$ with $\gamma:=(\rho, \beta)$ and $d_\gamma = 2$. As a reminder, one ending value of $\rho$= 0 enables the pure DT instability while the other ending value of $\rho=1$ active the pure DL instability.

In this study, three models—pFNO, pFNO*, and pCNN—described in Section \ref{sec:OLmethods} are employed to learn the two-parameter dependent operator $\hat{\mathcal{G}}_{(\rho,\beta)}$. As explained in the last paragraph in Section \ref{sec:OLmethods}, pFNO* is a simple variant of the baseline FNO method \cite{FNO2020} that includes the parameters in the codomain of the input function. On the other hand, pCNN has shown poor performance in learning the full operator $\hat{\mathcal{G}}_{(\rho,\beta)}$, with a high training error exceeding 3 percent, see Table \ref{Table1}. Therefore, we resort to two slightly restricted models (pCNN10 and pCNN40) which learn the single parameter ($\rho$) dependent operators $\hat{\mathcal{G}}_{(\rho,\beta=10)}$ and $\hat{\mathcal{G}}_{(\rho,\beta=40)}$, with each model being trained using one third of total dataset at $\rho$=10 and 40 respectively.

The learned operator at given parameters is expected to make recurrent predictions of solutions over an extended period. The training for such operators aims not only for accurate short-term predictions but also for robust predictions of long-term solutions with statistics similar to the ground truth. As demonstrated in previous studies \cite{Yu2023,YuH2024}, achieving this involves organizing the training data in a 1-to-20 pair, as expressed in Eq. \eqref{eq:opt_1-to-n}, optimized for accurately predicting 20 successive steps of outputs from a single input over a range of parameter values.

Table \ref{Table1} presents the relative training/validation errors for various models. The validation errors in Table \ref{Table1} are consistent with those reported in our previous work \cite{Yu2023,YuH2024}. Additional details on training and model hyper-parameters are provided in Appendix \ref{app:nn_detail}.

Fig. \ref{fig:disp_MKS_beta10} compares two randomly initialized sequences of front displacements predicted by two models (pFNO* and pCNN10) against the reference solutions at $\rho=[0,1/4,1/2,3/4,1]$ and $\beta$=10. A similar comparison for pFNO* and pCNN40 at $\beta=40$ is shown in Fig. \ref{fig:disp_MKS_beta40}. Additionally, Figs. \ref{fig:uSlope_MKS_beta10} and \ref{fig:uSlope_MKS_beta40} depict similar comparisons for the predicted front slope ($\disp_x$) at $\beta$=10 and 40, respectively.

All relevant model predictions at all fifteen parametric configurations $(\beta,\rho)\in [10,25,40]\times[0,1/4,1/2,3/4,1]$ are compared in Fig. \ref{fig:len_MKS} for the normalized total front length ($\int (\disp_x^2 +1 )^{1/2} dx /(2\pi) $), in Fig. \ref{fig:err_MKS} for the model errors accumulated through recurrent predictions, and in Fig. \ref{fig:corr_MKS} for the long-term auto-correlation function. This auto-correlation function characterizes  the long-term recurrently predicted solutions:
\begin{equation}
\mathcal{R} (r)   = \mathbb{E} \left(
 \int_\D \phi^*(x) \phi^*(x-r) dx 
 /\int_\D \phi^*(x) \phi^*(x) dx 
 \right).
\label{eq:corr}
\end{equation}
where $\phi^*(x)$ denotes the predicted solutions obtained after a sufficiently long time. Numerical calculation for the expectation $\mathbb{E}$ in Eq. \eqref{eq:corr} is implemented by averaging over seven randomly initialized sequences of model predictions for a time duration $1000<t/\Delta_t<4000$. Moreover, for each of the learned models $\G_{\theta,(\beta,\rho)} \approx \hat{\G}_{(\beta,\rho)} $, we compute an approximated dispersion relation:
\begin{equation}
\omega'(\kappa) = \log(J( \kappa,\kappa) ) /\Delta_t
\label{eq:dRel_model}
\end{equation}
where $J$ is the operator Jacobian 
\begin{equation}
 J(\kappa, \overline{\kappa} )
= 
\frac{\partial 
} 
{ \partial  \epsilon_\kappa  }  
\left|
 \mathcal{F}_{ \overline{\kappa} } 
    \{ 
       \G_{\theta,(\beta,\rho)}( 2\epsilon_\kappa \cos(\kappa x)    ) 
    \}  
\right|
,
\;\; 
\text{with } \kappa,\overline{\kappa} = 0, 1,2,...
\label{eq:Operator_Jacobian}
\end{equation}
This Jacobian is computed using the automatic differential tool (e.g., torch.autograd. functional.jacobian in PyTorch). Fig. \ref{fig:dRel_model} compares the dispersion relations by all models with the reference ones. Additionally, Fig. \ref{fig:dRel_model} shows one example of learned operator Jacobian, which is clearly diagonal dominant.

\subsection{Findings}

\noindent\textbf{Overall learning:}  
Our study underscores the robust learning capabilities of pFNO and pCNN methodologies in capturing the nuanced dynamics of flame front evolution, modulated by varying DL and DT instabilities blends. Both pFNO and pFNO* demonstrate good performance in learning the full two-parameter front evolution operator $\hat{\mathcal{G}}_{\rho,\beta}$ modulated by a $\rho$-varying blends of DL/DT instabilities as well as by a $\beta$-varying size of the largest unstable wavelength. 
While pCNN encounters difficulty in learning the full operator, the method still performs well in learning differently-hybrid instabilities when being restricted for the single-parameter operators $\hat{\mathcal{G}}_{\rho,\beta=10}$ and $\hat{\mathcal{G}}_{\rho,\beta=40}$.

\begin{figure*}
	\centerline{
	\includegraphics[width=1\linewidth]{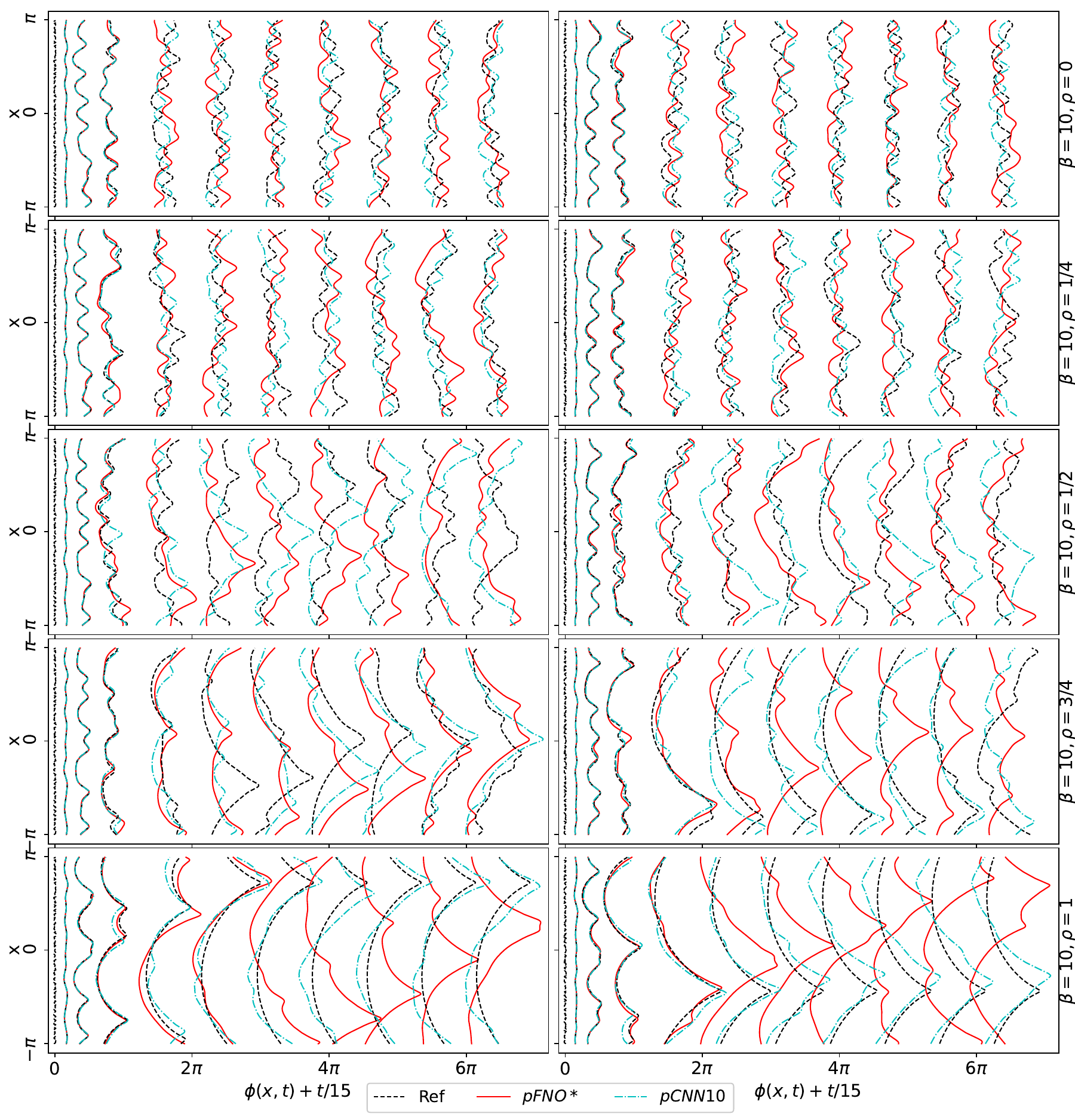}
	}
	\caption
	{
		\label{fig:disp_MKS_beta10}
Long-term solutions of flame front displacement $\disp(x,t)$ at $\beta=10 $ and $\rho = [0,1/4,1/2,3/4,1]$ (from top to bottom row). Black reference solutions to Eq. \eqref{eq:MKS} obtained using high-order numerical methods are compared against predictions by the operator-learning methods of pFNO* (red) and pCNN (cyan). The left and right columns correspond to two randomly initialized solution sequences, each showing eleven snapshots of $\disp(x,t)$ at $t/0.15$ = 0, 50, 125, 250, 500, 750, 1000, 1250, 1500, 1750 and 2000. A time shift ($t/15$) is added to the displayed fronts to avoid overlap.
	}
\end{figure*}

\begin{figure*}
	\centerline{
	\includegraphics[width=1\linewidth]{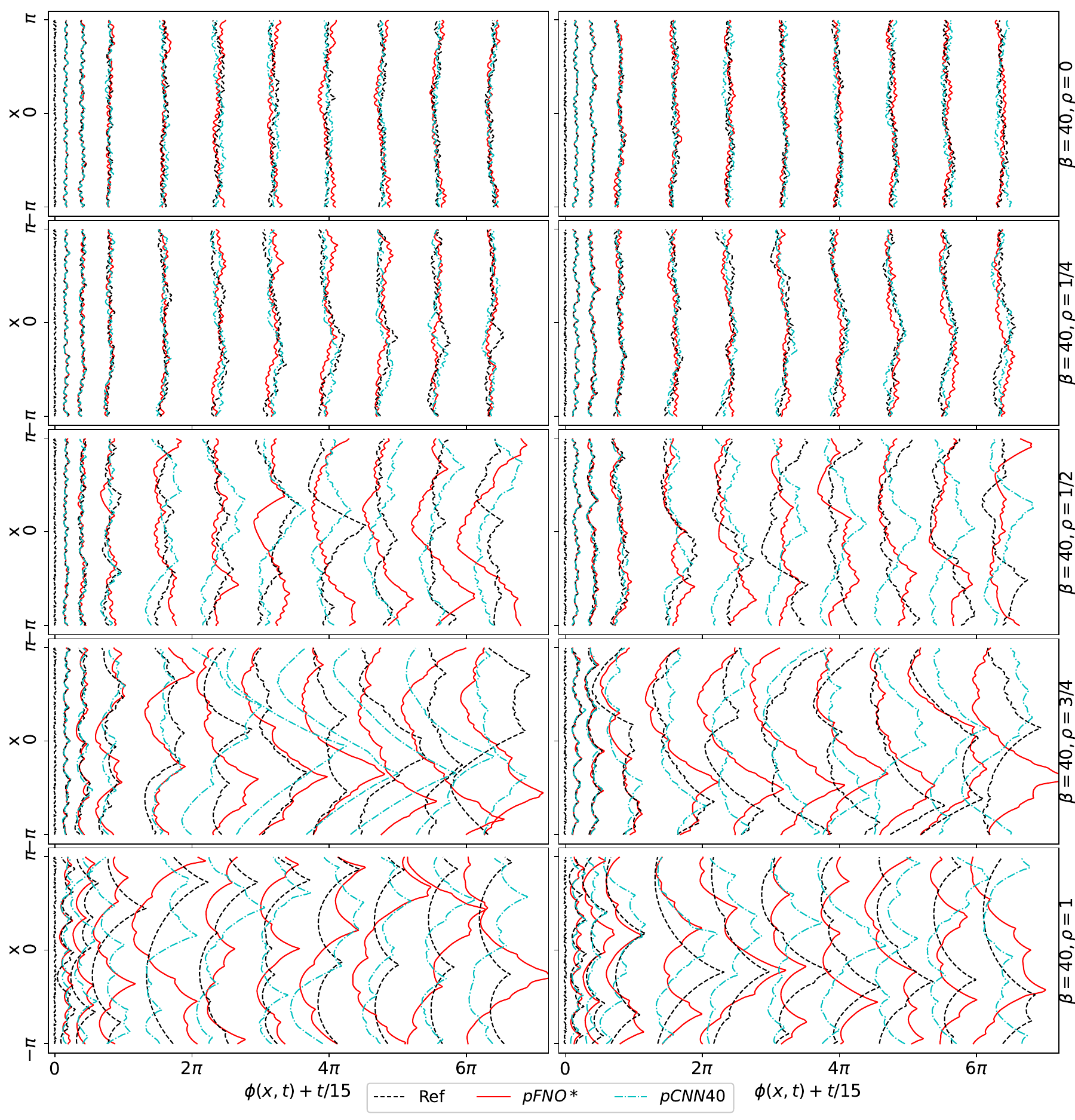}
	}
	\caption
	{
Comparison of flame front displacement predicted by pFNO* and pCNN40 at $\beta=40$. Other details are the same as in Fig. \ref{fig:disp_MKS_beta10}. 
		\label{fig:disp_MKS_beta40}
	}
\end{figure*}

\begin{figure*}
	\centerline{
		\includegraphics[width=1\linewidth]{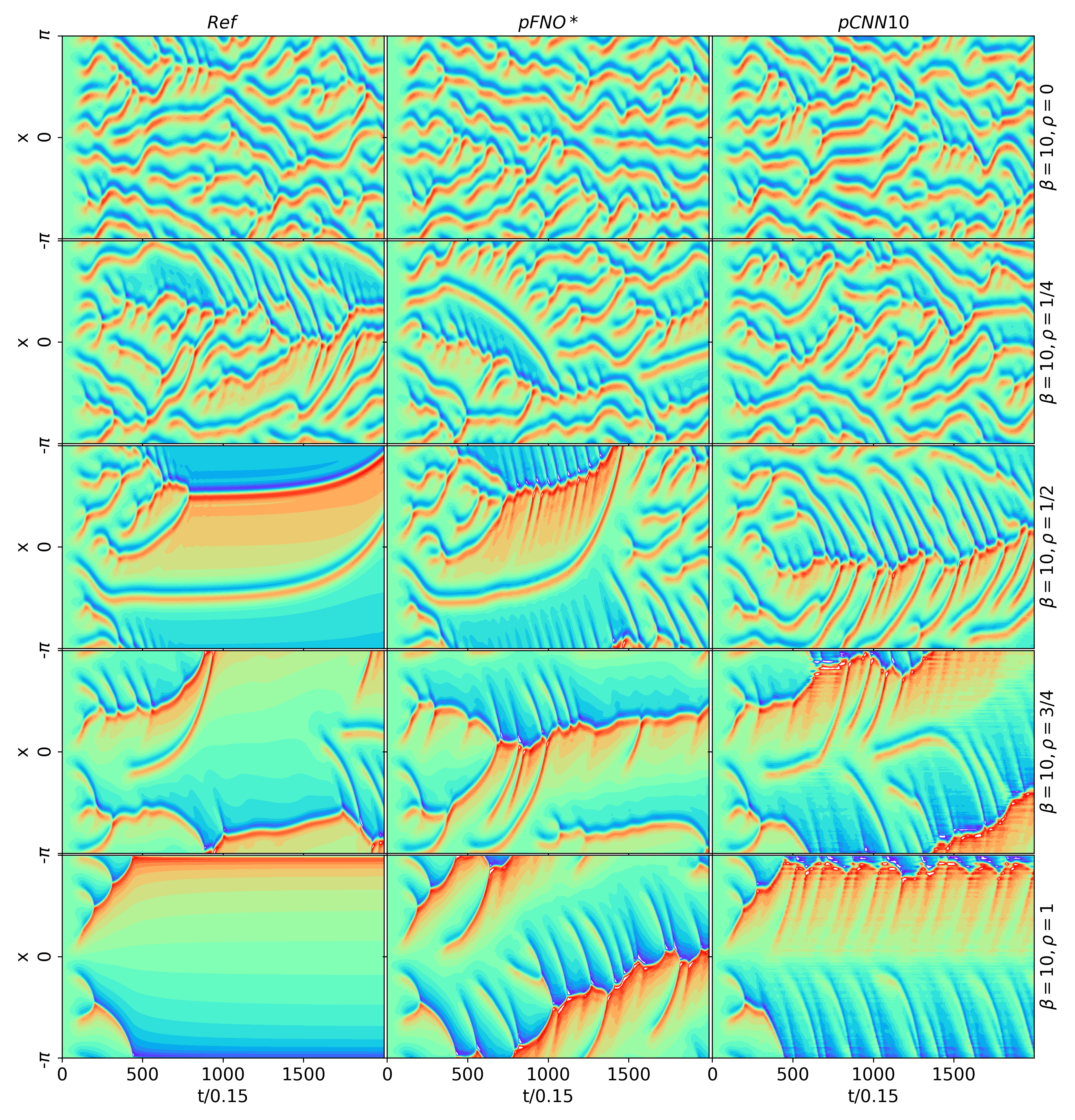}	   
}
	\caption{
		\label{fig:uSlope_MKS_beta10}
Comparison of front slope $\disp_x(x,t)$ calculated from a single instance reference solution (1st column) of Eq. \eqref{eq:MS} at $\beta=10$ and $\rho=[0,1/4,1/2,3/4,1]$, against predictions by pFNO* and pCNN (last two columns). Rainbow color indicates values from negative to positive.
	}
\end{figure*}

\begin{figure*}
	\centerline{
		\includegraphics[width=1\linewidth]{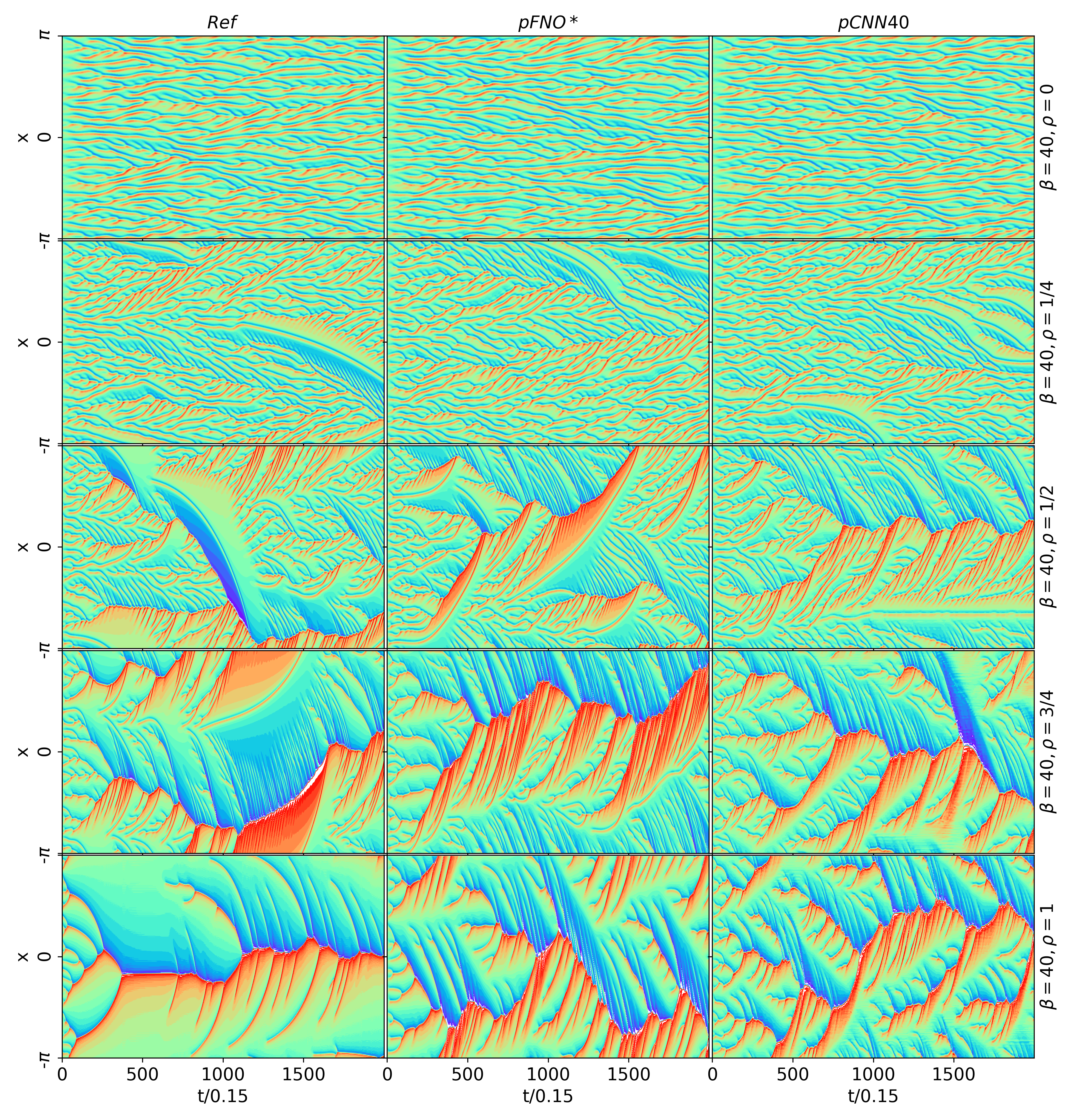}	   
}
	\caption{
		\label{fig:uSlope_MKS_beta40}
Comparison of front slopes predicted by pFNO* and pCNN40 at $\beta=40$. Other details are the same as in Fig. \ref{fig:uSlope_MKS_beta10}.
	}
\end{figure*}

\begin{figure*}
	\centerline{
		\includegraphics[width=1\linewidth]{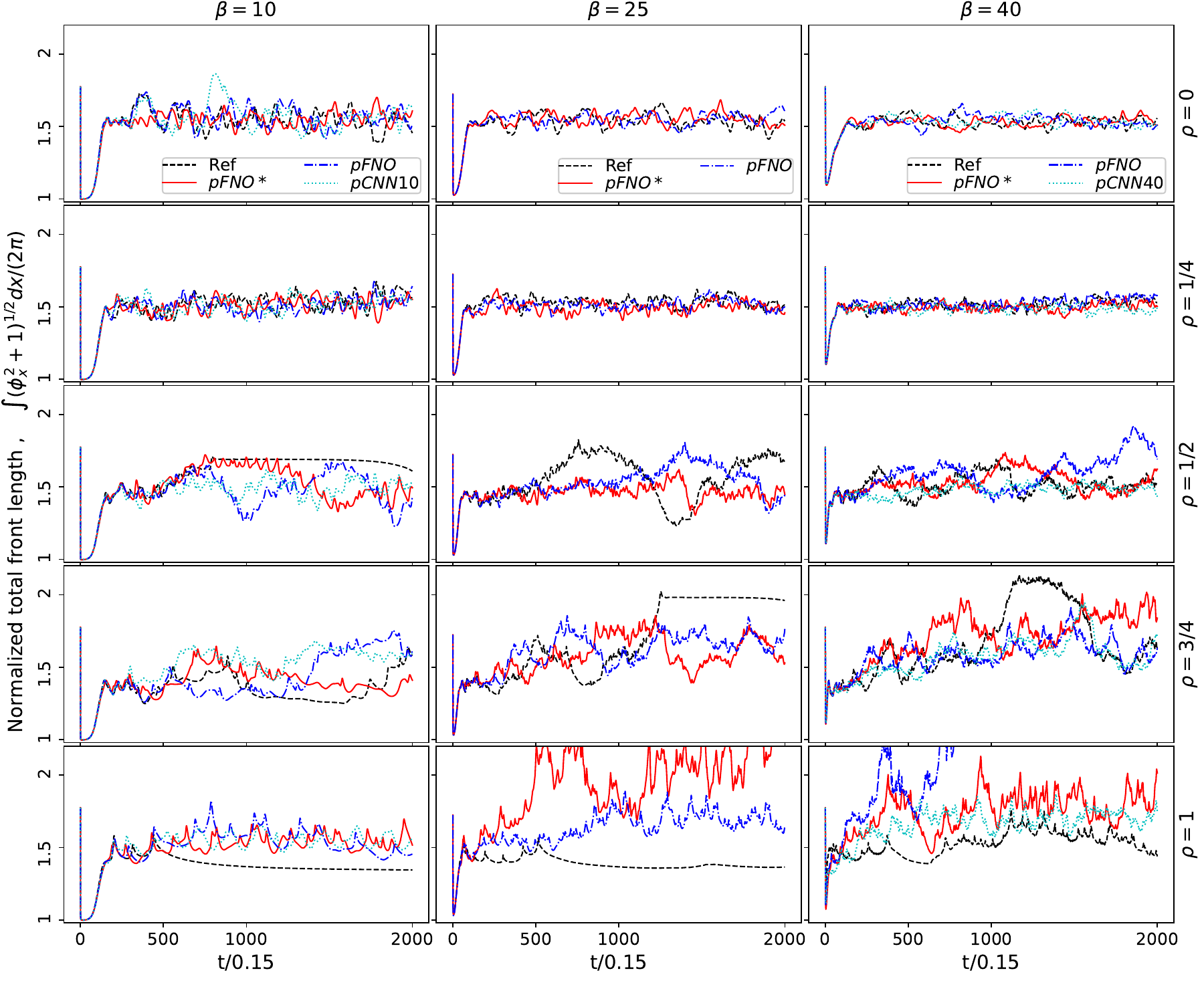}
   }
	\caption{
		\label{fig:len_MKS}
Normalized total front length comparison over 15 parametric configurations of $(\beta,\rho)=[10,25,40]\times[0,1/4,1/2,3/4,1]$. The black curve represents a single instance of reference solutions obtained from Eq. \eqref{eq:MKS}. Predictions by pFNO* are shown in red, pFNO in blue, and pCNN10/pCNN40 in cyan.
	}
\end{figure*}

\begin{figure*}
	\centerline{
		\includegraphics[width=1\linewidth]{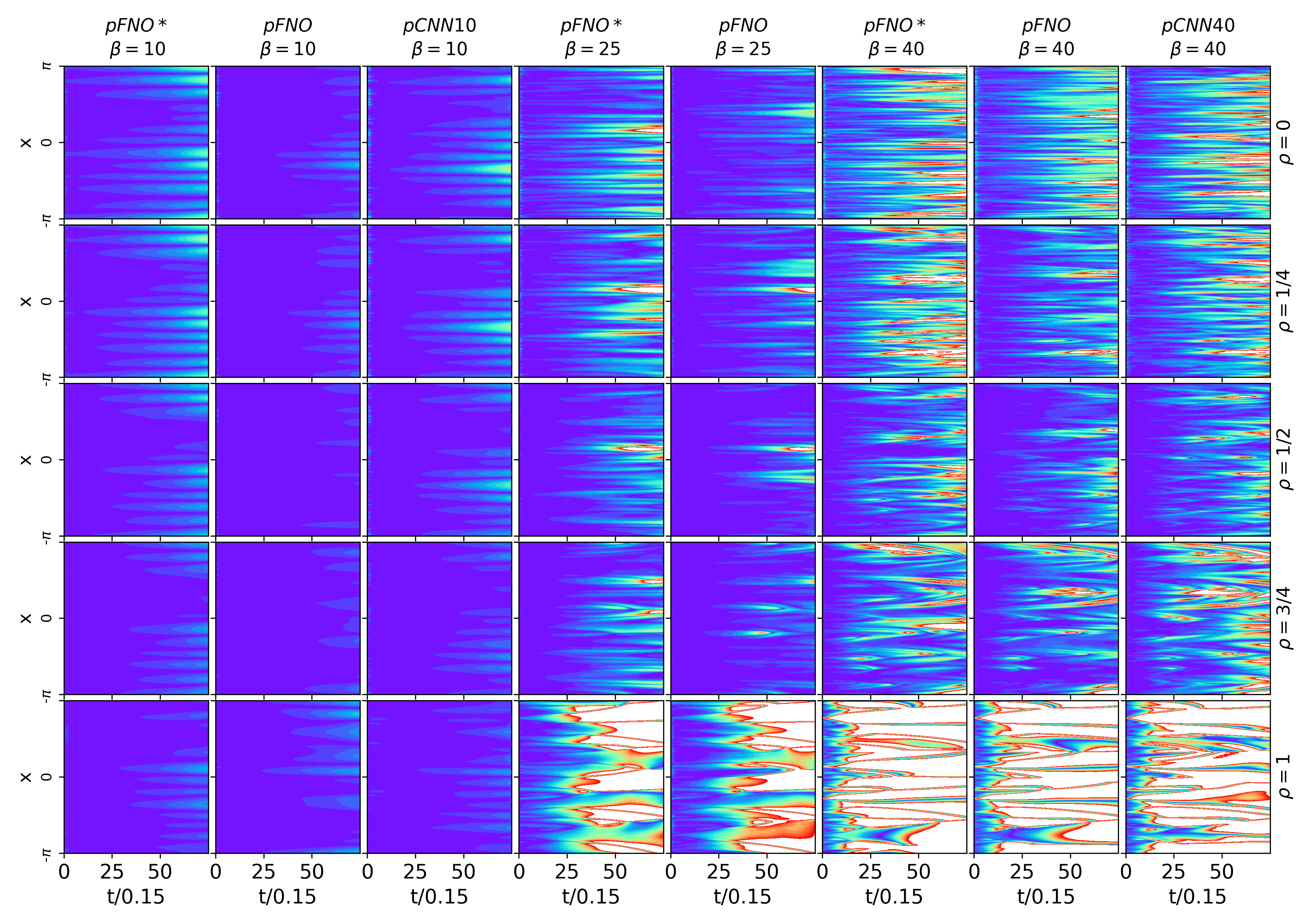}
   }
	\caption{
		\label{fig:err_MKS}
Time evolution of the relative $L2$-error between the reference solution of Eq. \eqref{eq:MKS} and the predicted solutions by pFNO*, pFNO, and pCNN10/pCNN40. The reference solutions are initialized with random conditions at 15 parametric configurations of $(\beta,\rho)=[10,25,40]\times[0,1/4,1/2,3/4,1]$. Rainbow colors from blue to red represent values ranging from 0 to 0.1; values above 0.1 are truncated and displayed as white.
	}
\end{figure*}

\noindent\textbf{Short term learning:} 
Across the board, all learned models (pFNO,pFNO* and pCNN10/40) demonstrate good accuracy in short-term predictions, with training/validation errors below 2 percent (Table \ref{Table1}) and small accumulated errors (Fig. \ref{fig:err_MKS}). This precision extends to various metrics, including front displacement, front slope, and normalized front length (Fig. \ref{fig:disp_MKS_beta10}, \ref{fig:disp_MKS_beta40}, \ref{fig:uSlope_MKS_beta10}, \ref{fig:uSlope_MKS_beta40}, \ref{fig:len_MKS} for $t \leq 50\Delta_t$), affirming the models' fidelity in capturing short-term dynamics. Moreover, pFNO demonstrates the smallest error and is the most accurate model for learning short term solutions. 

\noindent\textbf{Long term learning:} 
Detailed analysis of reference solutions unveils distinct characteristics of isolated instabilities. At $\rho=1$, DL fronts evolve toward a single giant cusp structure, either remaining stationary at small $\beta=10$ (Fig. \ref{fig:disp_MKS_beta10}) or exhibiting noise-induced wrinkles at larger $\beta=40$ (Fig. \ref{fig:disp_MKS_beta40}). Conversely, at $\rho=0$, DT fronts adopt an overall flat shape interspersed with oscillatory wavy structures, with decreasing wavelength and amplitude as $\beta$ increases from 10 to 40 (Figs. \ref{fig:disp_MKS_beta10}, \ref{fig:disp_MKS_beta40}). The slope plots for DT front evolution result in the typical zebra stripe pattern (Figs. \ref{fig:uSlope_MKS_beta10}, \ref{fig:uSlope_MKS_beta40}). Intermediate values of $\rho$ showcase a gradual transition between these features, with the front structure blending wavy oscillations together with the cusp shape.

\begin{figure*}
	\centerline{
		\includegraphics[width=1.15\linewidth]{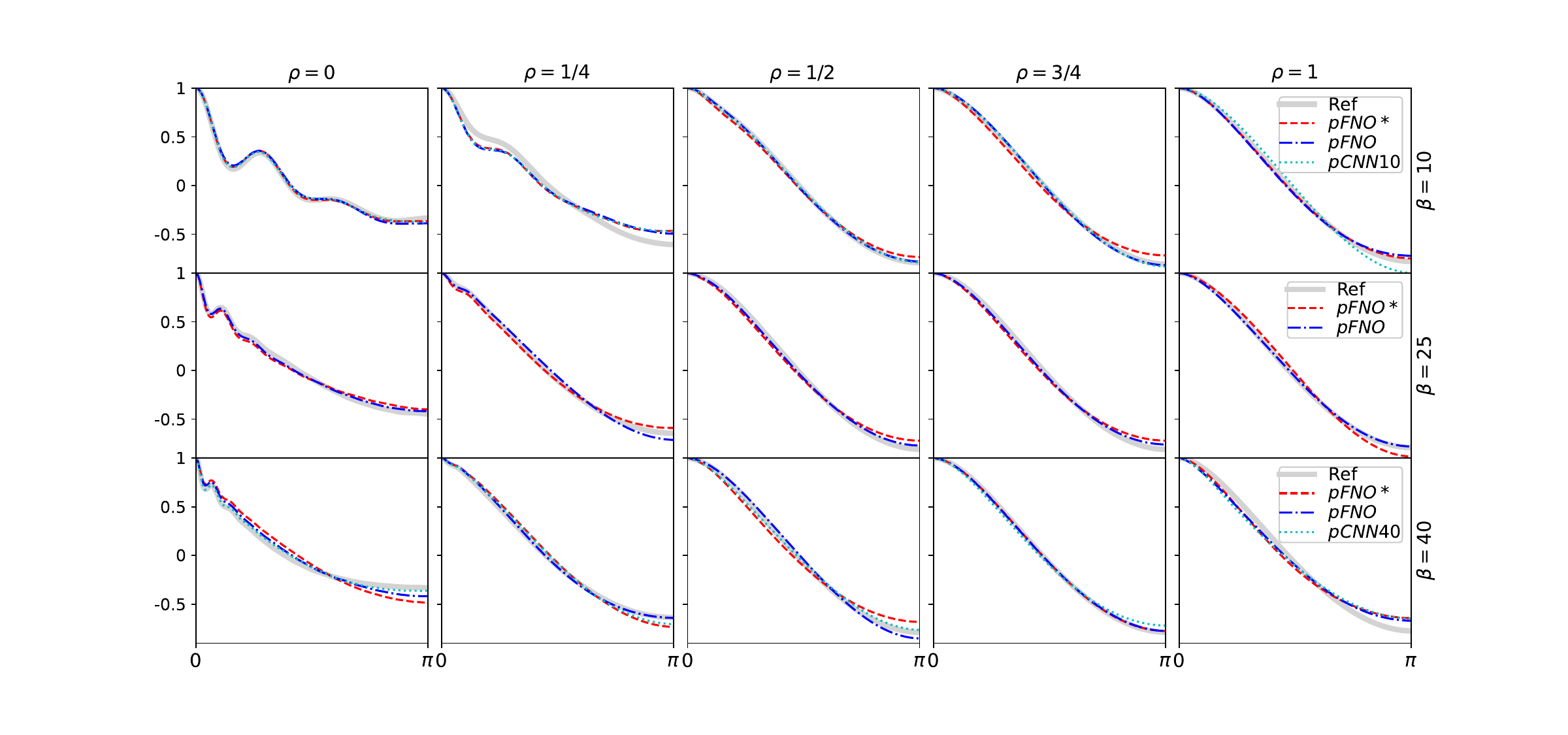}
   }
	\caption{
		\label{fig:corr_MKS}
The auto-correlation functions (Eq. \ref{eq:corr}) calculated from the long-term reference solutions at 15 parametric configurations of $(\beta,\rho)=[10,25,40]\times[0,1/4,1/2,3/4,1]$ are compared against those computed from the long-term predictions learned using pFNO*, pFNO, and pCNN10/pCNN40.
	}
\end{figure*}

Long-term predictions by all learned models (pFNO*, pFNO, pCNN10/pCNN40) accurately replicate these characteristic behaviors across diverse parametric configurations, encompassing pure DL and DT instabilities as well as blended scenarios. Quantitative comparisons through auto-correlation functions (Fig. \ref{fig:corr_MKS}) and total front length (Fig. \ref{fig:len_MKS}) confirm the models' proficiency in capturing long-term solutions.

\noindent\textbf{Learning challenges:} 
However, a common challenge across both pFNO and pCNN models lies in over-predicting the impact of noise-induced wrinkles, particularly noticeable at small $\beta=10$ (Fig. \ref{fig:disp_MKS_beta10}, \ref{fig:uSlope_MKS_beta10}). This tendency leads to an overestimation of the total front area, especially pronounced at lower $\beta$ values of 10 and 25 (Fig. \ref{fig:len_MKS} at $\rho=1$). When learning for the hybrid DL and DT instabilities, excessive noisy wrinkles also show up in all the model predictions (at $\rho=3/4$ in figs. \ref{fig:uSlope_MKS_beta10} and \ref{fig:uSlope_MKS_beta40}), however, the issue becomes less discernible toward smaller values of $\rho$ when DT instability plays a larger role, as also evident by the front length in Fig. \ref{fig:len_MKS}.

\begin{figure*}
	\centerline{
	\includegraphics[width=1\linewidth]{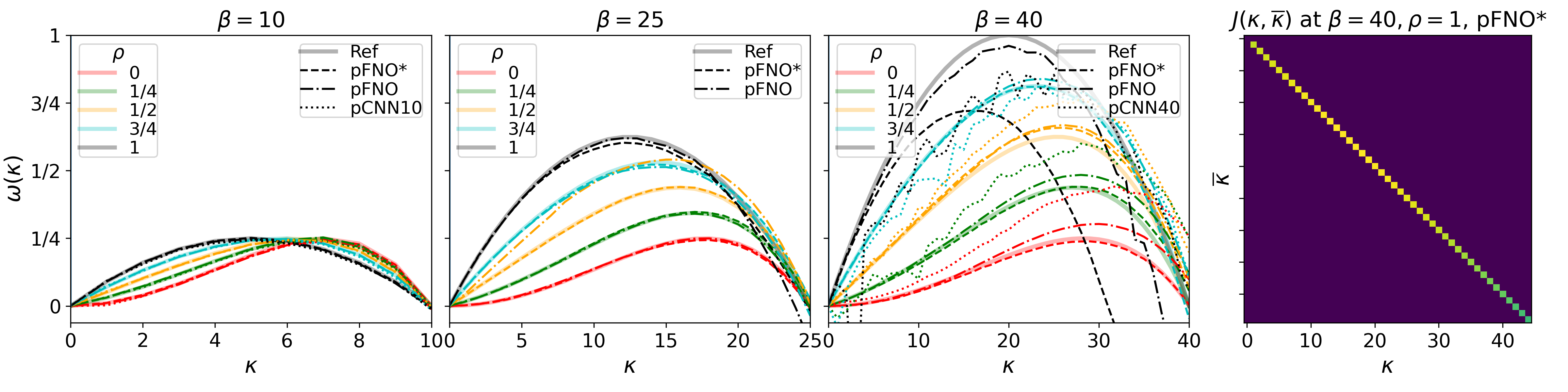}
   }
	\caption{
		\label{fig:dRel_model}
Left three columns: Comparison of reference dispersion relations (Eq. \ref{eq:dRel_ana}, solid lines) with those computed for the learned operators of all models (Eq. \ref{eq:dRel_model}, non-solid lines), where line colors indicate different $\rho$.
Right column: Illustration of a learned operator Jacobian (Eq. \ref{eq:Operator_Jacobian}), with dark colors indicating small values.
	}
\end{figure*}

\noindent\textbf{Extra finding:} 
It is particularly interesting to point out that the two models pFNO and pFNO* learn well on the parametric-dependent linear dispersion relations, as seen in Fig. \ref{fig:dRel_model}. Except for a moderate level of mismatch at a few parameter conditions toward large $\rho$ at 25 and 40, pFNO and PFNO* reproduce the relations quite accurately.

Such learning performance is impressive considering the fact that the data effective for learning these linear relations (i.e., the initial near-zero solutions) is just a tiny portion of the total dataset. For pCNN-based models, Fig. \ref{fig:dRel_model} shows pCNN10 learns the dispersion quite accurately while pCNN40 learned relations show a more significant deviation than ones by pFNO.

\section{Summary and Conclusion \label{sec:conclusion} }

This paper delves into the potential of machine learning (ML) for understanding and predicting the behavior of flames experiencing hybrid instabilities. These instabilities arise from the interplay of two key mechanisms: the Darrieus-Landau (DL) instability, driven by density gradients across a flame, and the Diffusive-Thermal (DT) instability, caused by heat and mass diffusion disparities.

The nonlinear development of unstable flames can be modeled by a well-known partial differential equation (PDE), specifically the Sivashinsky equation. By re-expressing the Sivashinsky equation, we introduce two parameters: $\rho$ and $\beta$. These parameters control the blending of DT and DL instabilities, as well as the cutoff wavelength for unstable flame behavior.

Our learning problem focuses on understanding the PDE solution time advancement operator under different parameter combinations. This operator, when repeatedly applied with its input solution as the output from the previous iteration, yields a time sequence of solutions of arbitrary length.
We employ two recently developed operator-learning models: parameterized Fourier Neural Operators (pFNO) and Convolutional Neural Networks (pCNN). Our findings demonstrate that both pFNO and pCNN models effectively capture the intricate flame dynamics under varying DT/DL instabilities (due to $\rho$ variations). Specifically:

\noindent\textbf{Short-Term Predictions:} 
 All learned models accurately predict short-term solutions and dispersion relations.

\noindent\textbf{Long-Term Behavior: } 
The models also reproduce correct statistics, quantified by autocorrelation functions and total front length.

\noindent\textbf{pFNO Superiority:} Notably, pFNO outperforms pCNN by allowing the learning of the full two-parameter operator, enabling variation in both $\rho$ and $\beta$.

\noindent\textbf{Challenges:}   However, both pCNN and pFNO tend to overestimate noise-induced wrinkles associated with DL instability, leading to inaccurate predictions of the total flame area, especially at lower instability levels.

In conclusion, this work showcases the potential of operator learning methodologies for analyzing complex flame dynamics arising from hybrid instabilities. While challenges persist, particularly related to noise overestimation, these methods offer promising tools for understanding and predicting real-world flame behavior in combustion systems \cite{hodzic2017a,hodzic2017b,hodzic2018,YBL15,YU2017nonlinear,YNBL2019_CF,YL2019_Equation,YNCL2020_JFM}. Future research directions may involve incorporating additional physical mechanisms or exploring alternative learning architectures to further enhance the accuracy and robustness of these models.

\begin{acknowledgments}
The authors gratefully acknowledge the financial support by the Swedish Research Council (VR-2019-05648). 
The computations were enabled by resources provided by the National Academic Infrastructure for Supercomputing in Sweden (NAISS), at ALVIS and Tetralith, partially funded by the Swedish Research Council through grant agreement no. 2022-06725.
\end{acknowledgments}

\section*{Data Availability Statement}
The code and data that support the findings of this study are openly available at www.github.com/RixinYu/ML\_paraFlame.

\section*{Declaration of Interest}
The authors have no conflicts to disclose.


\appendix
\section{ Model hyper-parameters and training details  \label{app:nn_detail} }
All models undergo training for 1000 epochs with a batch size of 1000, employing the Adam optimizer with a learning rate set at 0.0025 and a weight decay of 0.0001. A scheduler step size of 100 and a gamma value of 0.5 are applied for learning rate adjustment. To stabilize the training process, the maximum norm of gradients is clipped above 50.

Training for pFNO and pFNO* is conducted over the entire dataset using a single GPU (NVIDIA Tesla A40), taking approximately 28 and 47 hours, respectively. In contrast, training pCNN10 and pCNN40 is performed on one-third of the training dataset, with both models requiring around 38 hours using a single GPU. Conversely, training the pCNN model (to learn the full two-parameters operator) using the entire dataset takes 26 hours utilizing four GPUs.

The pFNO and pFNO* networks are configured with $L=4$ levels and $d_{\z}=30$ channels, with two hyperparameters set: $\kk_\text{max}=128$ and $N_\gamma=5$. To reduce model size, all pFNO methods share most of the trainable parameters within a single parametric Fourier layer (Eq. \eqref{eq:pFNO}) across all layers $l = 0, ..., L-1$, except for those used to parameterize the function $D_l(\gamma)$. 

\bibliography{ReferencesParaDLDT}
\end{document}